\documentclass{article}
\usepackage{arxiv}

\usepackage[utf8]{inputenc} 
\usepackage[T1]{fontenc}    
\usepackage{hyperref}       
\usepackage{url}            
\usepackage{booktabs}       
\usepackage{amsfonts}       
\usepackage{nicefrac}       
\usepackage{microtype}      
\usepackage{lipsum}
\usepackage{graphicx}
\graphicspath{ {./images/} }
\usepackage{subfig}
\usepackage{comment}
\usepackage{subcaption}
\usepackage{wrapfig}
\usepackage{graphicx}%
\usepackage{multirow}%
\usepackage{multicol}
\usepackage{float}
\usepackage{makecell}
\usepackage{amsmath,amssymb,amsfonts}%
\usepackage{amsthm}%
\usepackage{array}
\usepackage{mathrsfs}%
\usepackage[title]{appendix}%
\usepackage{xcolor}%
\usepackage{textcomp}%
\usepackage{manyfoot}%
\usepackage{booktabs}
\usepackage{tabularx}
\usepackage{colortbl}
\arrayrulecolor{gray!60}
\arrayrulecolor{black} 
\usepackage{algorithm}%
\usepackage{algorithmicx}%
\usepackage{algpseudocode}%
\usepackage{listings}%
\usepackage{siunitx}
\usepackage{hyperref}

\title{VRWKV-Editor: Reducing Quadratic Complexity in Transformer-Based Video Editing}

\author{
 Abdelilah Aitrouga \\
  International Artificial Intelligence Center of Morocco\\
  University Mohammed VI Polytechnic\\
  Rabat, Morocco \\
  \texttt{abdelilah.aitrouga@um6p.ma} \\
   \And
 Youssef Hmamouche \\
  International Artificial Intelligence Center of Morocco\\
  University Mohammed VI Polytechnic\\
  Rabat, Morocco \\
  \texttt{youssef.hmamouche@um6p.ma} \\
  \And
 Amal El Fallah Seghrouchni \\
  International Artificial Intelligence Center of Morocco\\
  University Mohammed VI Polytechnic\\
  Sorbonne University, LIP6 - UMR 7606 CNRS, France\\
  Rabat, Morocco  \\
  \texttt{amal.elfallah-seghrouchni@um6p.ma} \\
}

\begin{document}
\maketitle
\newcommand\preprintnotice{%
  \begingroup
  \renewcommand\thefootnote{}\footnote{%
The Version of Record of this article is published in \textit{Multimedia Systems}, and is available online at \url{https://doi.org/10.1007/s00530-026-02649-4}%
  }%
  \addtocounter{footnote}{-1}%
  \endgroup
}
\preprintnotice
\begin{abstract}
In light of recent progress in video editing, deep learning models focusing on both spatial and temporal dependencies have emerged as the primary method. However, these models suffer from the quadratic computational complexity of traditional attention mechanisms, making them difficult to adapt to long-duration and high-resolution videos. This limitation restricts their applicability in practical contexts such as real-time video processing. To tackle this challenge, we introduce a method to reduce both the time and space complexity of these systems by proposing \texttt{VRWKV-Editor}, a novel video editing model that integrates a linear spatio-temporal aggregation module into video-based diffusion models. \texttt{VRWKV-Editor} leverages bidirectional weighted key-value recurrence mechanism of the \texttt{RWKV} transformer to capture global dependencies while preserving temporal coherence, achieving linear complexity without sacrificing quality. Empirical evaluations indicate that the proposed method achieves up to 3.7$\times$ speedup and over 70\% lower memory consumption than transformer-based attention approaches, while maintaining competitive performance in frame consistency and text alignment. Furthermore, a comparative study we conducted on videos with different sequence lengths confirms that the gap in editing speed between our approach and existing architectures becomes more significant with long videos. The source code and video demos for our model are available at \href{https://abdo-rg.github.io/VRWKV-Editor/}{\texttt{VRWKV-Editor}}.
\end{abstract}

\begin{figure*}[tbp]
    \centering
\includegraphics[width=\textwidth]{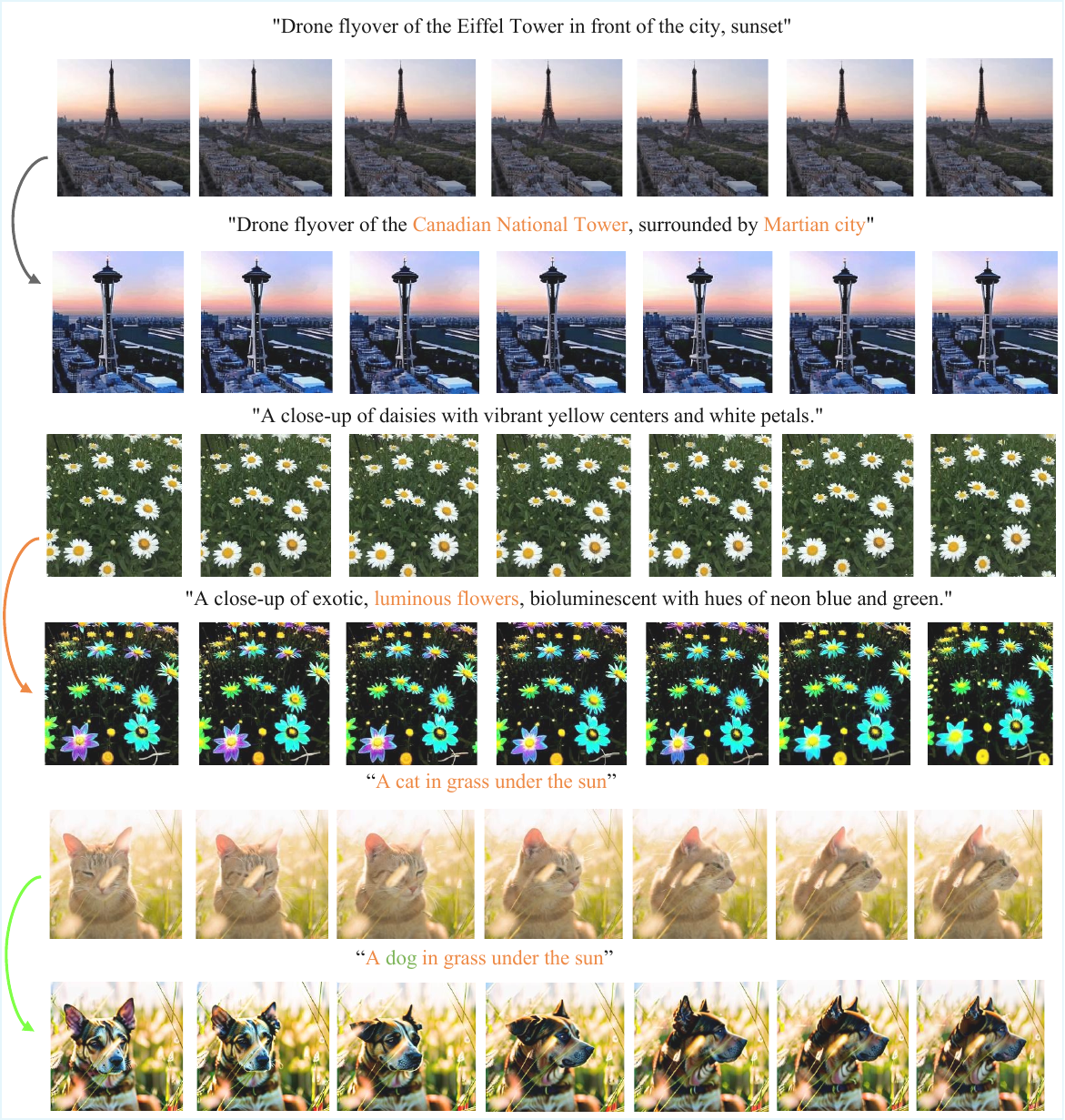}
\caption{Representative results produced by \texttt{VRWKV-Editor}. The proposed method is capable of object replacement, background modification, and style transformation, while consistently preserving the motion dynamics and visual characteristics of the original video. Additional qualitative examples are provided on the project page: \href{https://abdo-rg.github.io/VRWKV-Editor/}{\texttt{VRWKV-Editor}}.
}
    \label{fig0}
\end{figure*}


\section{Introduction}
The Transformer architecture has emerged as a dominant paradigm, revolutionizing fields from Natural Language Processing (NLP) to Computer Vision. Central to its effectiveness is the self-attention mechanism, which   captures the long-range dependencies within a sequence. However, a significant limitation of traditional attention is its computational and memory complexity, which scales quadratically with the input sequence length. This presents a critical bottleneck for applications with limited computational resources involving high-resolution data or real-time processing constraints.

In visual tasks, Vision Transformers (ViTs) \cite{dosovitskiy2020image} and their extensions \cite{liu2021swin,zhang2023vitaev2}, while demonstrating remarkable performance in image understanding, also inherit this problem and suffer from quadratic computational complexity. Videos are characterized by strong spatiotemporal dependencies, which distinguishes them from collections of independent images. Consequently, models must ensure consistency within and across frames. This dual requirement leads to an additional increase in model's complexity.

Within the framework of text-driven video editing, optimized architectures such as Swin-transformer \cite{liu2021swin} also presents the quadratic computational bottleneck that limits scalability despite its effectiveness in handling local spatial relationships. Reducing this complexity is thus critical for both research and industry, as efficient models lower costs and enable real-time deployment in consumer applications such as mobile video editors, while also allowing researchers to experiment and iterate without relying on large-scale computational resources.

Concurrently, modern RNN-like architectures in NLP  have emerged as efficient solutions for processing lengthy sequences linearly  while preserving the parallelization advantages of transformers. 
Illustratively, RWKV, Mamba, xLSTM, and Fast Transformer \cite{peng2023rwkv, gu2023mamba, beck2024xlstm, katharopoulos2020transformers} each employ a distinct mechanism to linearize the classical Transformer. However, these methods are fundamentally based on two main ideas: either using an aggregation mechanism with a linear kernel in place of the softmax function, or replacing the dot-product with element-wise multiplication as a means of linearization.

In accordance with this progress, \texttt{VRWKV} \cite{duan2024vision} extends \texttt{RWKV}  two-dimensional vision tasks such as image classification, detection and segmentation, demonstrating robust global processing capabilities while efficiently handling sparse inputs and maintaining linear computational complexity. Notably, \texttt{VRWKV} provides reduced spatial aggregation complexity and enables seamless processing of high-resolution images without requiring window operations.\\
Expanding on these insights, we present a new video editing framework, termed \textbf{\texttt{VRWKV-Editor}},
which leverages the linear complexity and global attention capabilities of this mechanism. Furthermore, rather than employing straightforward concatenation in the skip connections of U-Net architecture, we utilize bidirectional processing inspired by \texttt{VRWKV’s} Bi-WKV mechanism to merge feature maps from corresponding encoding paths with those from preceding decoding up-convolutional layers. This design enables the model to learn continuous motion patterns in input videos and transfer them to edited outputs according to user prompts, while significantly reducing computational complexity. Remarkably, our experimental results demonstrate that this system achieves computational efficiency superior to that of state-of-the-art methods, all while retaining comparable performance. In general, this work presents the following contributions:
\begin{itemize}
    \item[-] We propose \textbf{\texttt{VRWKV-Editor}}, a novel video editing framework that offers a cost-effective and efficient solution with reduced computational requirements. It preserves the ability to capture long-range dependencies and flexibly handle sparse inputs, while reducing complexity to a linear scale. This reduction removes the need for window operations when processing long video sequences, making \texttt{VRWKV-Editor} a more scalable and practical approach for video editing tasks.
    \item[-] We demonstrate significant improvements in computational efficiency through the integration of \texttt{VRWKV's} bidirectional attention mechanism, reducing model complexity without performance degradation.
    \item[-] We provide comprehensive experimental validation showing that linear attention mechanisms can effectively replace traditional quadratic attention in video understanding tasks while offering substantial computational advantages.
\end{itemize}

The remainder of this paper is organized as follows: Section \ref{sec:rw} reviews related work in video editing and linear attention mechanisms. Sections \ref{sec:bg} and \ref{sec:md} detail background concepts relevant to our study and the proposed \texttt{VRWKV-Editor} architecture with its key components. Section \ref{sec:exp} presents a comprehensive analysis and experimental results. In Section \ref{sec:lm}, we highlight potential limitations of this work. Finally, Section \ref{sec:cc} concludes with a discussion of the study's implications and potential directions for future research.

\section{Related works}\label{sec:rw}
This section discusses three research areas  related to our approach; video editing methods using diffusion models, which highlight the computational challenges inherent in current approaches; linear attention mechanisms, which offer efficiency gains over quadratic attention; and computational efficiency approaches in video processing, which directly motivate our integration of linear attention for scalable video editing.

\subsection{Video Editing with Diffusion Models}
Recent developments in diffusion model-based video editing have introduced various approaches to address the computational and quality challenges inherent in video processing \cite{sun2024diffusion,wang2023zero,wu2023tune,TU2026111966}. Early approaches focused on adapting image diffusion models for video tasks through temporal extensions. Tune-A-Video \cite{wu2023tune} pioneered the concept of one-shot tuning in video editing by transforming spatial self-attention layers into sparse-causal attention mechanisms. VidToMe \cite{li2024vidtome} leveraged temporal redundancy through Token Merging (ToMe) to enhance consistency while reducing computational overhead. However, these methods still rely on traditional attention mechanisms with quadratic complexity, limiting their scalability to high-resolution and long-sequence video processing. Methods like Prompt-to-Prompt (P2P) \cite{liu2023video} and Plug-and-Play (PnP) \cite{tumanyan2023plug} have demonstrated the effectiveness of attention feature injection techniques, where cross-attention and self-attention maps from reconstruction branches guide the editing process. TokenFlow \cite{geyer2023tokenflow} and FLATTEN \cite{cong2023flatten} utilize inter-frame feature similarity and optical flow-based correspondence respectively to maintain temporal consistency across video frames. While effective, these approaches face computational bottlenecks when processing high-resolution videos due to their reliance on traditional quadratic attention mechanisms. Recent works have incorporated structural guidance through depth maps, optical flow, and motion vectors to improve temporal consistency. FlowVid \cite{liang2024flowvid} and MoCA \cite{yan2023motion} integrate edited first frames with optical flow information to maintain motion fidelity, while approaches like ControlVideo \cite{zhang2023controlvideo} leverage multi-modal conditions for enhanced controllability.\\
Complementary to these editing approaches, advances in video generation architectures have explored pixel-based, latent-based, and hybrid designs \cite{liu2023video, ho2022video,CHEREL2024103866, zhang2025show}. Research has shown that pixel-based diffusion models excel in generating low-resolution videos with superior text-video alignment and motion fidelity, while latent-based models offer computational efficiency but may compromise alignment quality \cite{zhang2025show}. Hybrid approaches that combine pixel-based keyframe generation with latent-based super-resolution have demonstrated the potential to achieve both high quality and computational efficiency \cite{hong2022cogvideo}. Effective video models incorporate temporal layers within U-Net architectures, including temporal convolution layers and temporal attention mechanisms to facilitate dynamic temporal data assimilation \cite{shao2025learning}. The integration of spatial and temporal processing remains crucial for maintaining coherent motion patterns across video sequences \cite{meral2024motionflow}.

\subsection{Linear Attention Mechanisms in Vision}
The computational limitations of traditional attention mechanisms have sparked interest in linear attention alternatives that maintain performance while reducing complexity \cite{LI2024104166, CHEREL2024103866}. Recent advances in natural language processing have introduced models with linear feature aggregation mechanisms, such as RWKV and Mamba \cite{peng2023rwkv, gu2023mamba}, which emerge as efficient solutions for processing lengthy sequences while maintaining transformer-like capabilities. 
Complementary work has explored linear attention from kernel-based and low-rank approximation perspectives. Performer \cite{choromanski2021rethinking} employs FAVOR+ kernel projections to approximate softmax attention with linear complexity, enabling efficient global receptive fields in vision tasks. Linformer \cite{wang2020linformer} reduces attention cost via low-rank projections of key and value matrices, demonstrating that self-attention can be inherently low-dimensional in many vision backbones. RetNet \cite{sun2024retentive} further advances this direction by unifying recurrence and attention through retention mechanisms, offering a scalable alternative for long-context modeling in vision and multimodal workloads.
In the vision domain, these mechanisms have increasingly been adapted to high-resolution and generative modeling tasks. Diffusion-RWKV \cite{fei2024diffusion} shows that RWKV-based architectures can efficiently process patchified inputs while significantly reducing spatial aggregation costs, achieving competitive performance without windowing or cached-group operations. VRWKV \cite{duan2024vision} extends these principles to visual modeling, combining linear complexity with strong global context processing. Its bidirectional Bi-WKV mechanism supports comprehensive spatial–temporal reasoning, making it particularly effective for tasks such as video editing. Together, these works highlight the growing viability of linear attention as an efficient and scalable alternative to transformer-based vision architectures.

\subsection{Computational Efficiency in Video Processing}
The computational demands of video processing have motivated various efficiency-oriented approaches that directly relate to our linear attention integration \cite{XU2025104545,SUN2026112091}. Current video editing methods exhibit significant variations in computational requirements, with training-free methods like TokenFlow \cite{geyer2023tokenflow}, Rerender \cite{yang2023rerender}, and RAVE \cite{kara2024rave} showing superior efficiency while maintaining satisfactory performance. However, these methods still face scalability challenges when processing high-resolution videos or long sequences due to underlying quadratic attention mechanisms \cite{hong2022cogvideo, singer2022make, zhang2024towards}. The hierarchical nature of traditional attention mechanisms, while beneficial for capturing multi-scale features, often results in increased parameter counts and computational overhead, particularly challenging for real-time video processing applications \cite{bertasius2021space}. This limitation becomes especially pronounced in video editing scenarios where maintaining temporal consistency requires processing multiple frames simultaneously \cite{liu2023video,wu2023tune,tumanyan2023plug}.
Our work addresses these computational limitations by proposing \texttt{VRWKV-Editor}, which leverages the linear complexity of \texttt{VRWKV} to achieve superior computational efficiency while maintaining high-quality video editing performance. Unlike previous approaches that rely on quadratic attention mechanisms, our method enables seamless processing of high-resolution videos without requiring window operations or hierarchical attention patterns.

\section{Background}\label{sec:bg}
This section provides the necessary foundations that are used in our approach. First, we discuss latent variable refinement techniques that improve the efficiency and quality trade-off. We then review diffusion models in the context of video generation and highlight their extension from image-based frameworks to spatio-temporal domains, followed by a detailed overview of the \texttt{VRWKV} architecture, which offers a linear complexity alternative to traditional attention mechanisms. These components form the conceptual backbone for our proposed method.
\subsection{Diffusion Models and Latent Variable Refinement} Diffusion models represent \cite{ho2020denoising} a powerful class of generative models that operate through a progressive noise addition and removal process. Starting with an original image $x_0$, the forward diffusion process gradually corrupts the data by adding Gaussian noise over $T$ timesteps. At each step $t$, the noisy image $x_t$ is obtained by:

\begin{equation}
    x_t = \sqrt{\alpha_t} x_{t-1} + \sqrt{1-\alpha_t} \epsilon_t,
\end{equation}

where $\alpha_t$ represents a noise schedule parameter and $\epsilon_t \sim N(0,I)$ is Gaussian noise with the same dimensions as $x_0$.\\
The reverse process learns to denoise these corrupted images, effectively modeling the data distribution. For conditional generation tasks, we can extend this framework to model the conditional distribution $p_{\theta}(x_t|c,x_0)$, where $c$ represents conditioning information such as text prompts or class labels. The conditional model learns the following distribution:

\begin{equation}
    p_\theta(x_0 | c) = \int p_\theta(x_{0:T} | c) dx_{1:T},
\end{equation}

\begin{equation}
    p_\theta(x_{0:T} | c) = p(x_T) \prod_{t=1}^T p_\theta(x_{t-1} | x_t, c),
\end{equation}

where $x_T \sim N(0,I)$ represents pure noise. The model parameters $\theta$ are optimized by maximizing the variational lower bound of the log-likelihood:

\begin{equation}
    \mathcal{L} = \mathbb{E}_{x_0, c, t, \epsilon} \left[\|\epsilon - \epsilon_\theta(x_t, t, c)\|^2\right].
\end{equation}

The network $\epsilon_\theta$ learns to predict the noise added in each time step given the noisy image $x_t$, the time step $t$, and the conditioning information $c$.\\
To improve computational efficiency, latent diffusion models perform the diffusion process in a compressed latent space rather than directly in the pixel space. This approach utilizes a pre-trained autoencoder consisting of an encoder $E$ that maps images to latent representations $z = E(x)$ and a decoder $D$ that reconstructs images from latents.\\
The diffusion process then operates in this latent space, learning the distribution of encoded representations. This approach significantly reduces memory and computational costs while maintaining high-quality generation capabilities \cite{rombach2022high}. During inference, the model generates latent representations through the reverse diffusion process, which are then decoded back to pixel space using the pretrained decoder.\\
This framework enables efficient conditional image generation by combining the expressiveness of diffusion models with the computational advantages of latent space operations, making it particularly suitable for high-resolution image synthesis tasks.

\subsection{Diffusion Models for Video Processing}
Beyond their success in image generation, diffusion models have recently been extended to video synthesis, where they enable high-quality spatio-temporal generation and editing \cite{ZHAO2026111867, zhao2025motiondirector, wei2024dreamvideo}. These models operate by learning to reverse a gradual noise addition process, where data is progressively corrupted through a forward Markov chain and then restored through a learned reverse process.\\
The transition from image to video generation involves inflating 2D models to handle spatiotemporal dimensions, typically through pseudo-3D convolutions where 3×3 kernels are replaced by 1×3×3 kernels, and temporal attention layers are added to transformer blocks \cite{zhang2025show, molad2023dreamix}. However, this extension introduces significant computational challenges due to the quadratic complexity of attention mechanisms when processing multiple frames simultaneously.

\subsection{VRWKV Architecture}
\texttt{VRWKV} consists of three main components: Receptance $(R)$, Key $(K)$, and Value $(V)$ transformations, combined with a time-mixing mechanism specifically adapted for vision tasks. For a given input sequence of video frames $X_t \in \mathbb{R}^{f \times c \times h \times w}$ consisting of $f$ frames of size $h \times w$
and the number of channels $c$, the \texttt{VRWKV} operations are defined as:

\begin{align}
R_t = W_R \otimes (\mu_R \odot X_t + (1 - \mu_R) \odot X_{t-1}),\\
K_t = W_K \otimes (\mu_K \odot X_t + (1 - \mu_K) \odot X_{t-1}),\\
V_t = W_V \otimes (\mu_V \odot X_t + (1 - \mu_V) \odot X_{t-1}).
\label{eq5}
\end{align}

where $W_R$, $W_K$, $W_V$ are learnable projection matrices, $\mu_R$, $\mu_K$, $\mu_V$ are learnable interpolation factors and $\otimes$ denotes a spatiotemporal convolution, defined as a standard $2D$ spatial convolution applied on a temporally interpolated frame representation $\mu \odot X_t + (1-\mu)\odot X_{t-1}$. This differs from standard 3D convolution since it only considers the current and previous frame, rather than a full temporal kernel. \\
The core innovation of \texttt{VRWKV} lies in its weighted key-value mechanism that maintains temporal coherence:

\begin{equation}
\label{eq6}
w k v_t=\operatorname{Bi-WKV}(K, V)_t=\frac{\sum_{i=0, i \neq t}^{T-1} e^{-(|t-i|-1) / T \cdot w+k_i} v_i+e^{u+k_t} v_t}{\sum_{i=0, i \neq t}^{T-1} e^{-(|t-i|-1) / T \cdot w+k_i}+e^{u+k_t}}
\end{equation}

where $T$ represents the total number of tokens, $w$ and $u$ are two $d
$-dimensional learnable vectors that represent channel-wise spatial decay and the bonus indicating the current token, respectively. $k_t$ and $v_t$ denote $t$-th feature of $K$ and $V$.\\
The final output incorporates both the weighted key-value information and the receptance gate:
\begin{equation}
O_t = \sigma(R_t) \odot {wkv}_t,
\end{equation}

where $\sigma$ represents the sigmoid activation function.\\
The output $O_t$ from the time-mixing block is then processed through a channel-mixing block that enhances feature representation across different channels. The channel-mixing mechanism operates as follows:
\begin{align}
R_{c,t} &= W_{R_{c}} \otimes (\mu_{R_{c}} \odot O_t + (1 - \mu_{R_{c}}) \odot O_{t-1}),\\
K_{c,t} &= W_{K_{c}} \otimes (\mu_{K_{c}} \odot O_t + (1 - \mu_{K_{c}}) \odot O_{t-1}),
\end{align}

where $W_{R_{c}}$ and $W_{K_{c}}$ are channel-mixing projection matrices, and $\mu_{R_{c}}$, $\mu_{K_{c}}$ are learnable interpolation parameters for the channel-mixing component. The channel-mixed output is computed as:

\begin{equation}
CM_t = \sigma(R_{c,t}) \odot \max(K_{c,t} \cdot W_{c}, 0)^2,
\end{equation}

where $W_{c}$ is a learnable transformation matrix and the squared ReLU activation $\max(\cdot, 0)^2$ provides enhanced non-linearity to model channel interactions. The final layer output combines both time-mixing and channel-mixing components through a residual connection:

\begin{equation}
Y_t = X_t + O_t + CM_t.
\end{equation}

This dual-mixing architecture enables \texttt{VRWKV} to capture both temporal dependencies through the time-mixing block and cross-channel feature interactions through the channel-mixing block, providing comprehensive spatiotemporal modeling capabilities for vision tasks.
\begin{figure*}[htbp]
    \centering
\includegraphics[width=0.98\textwidth]{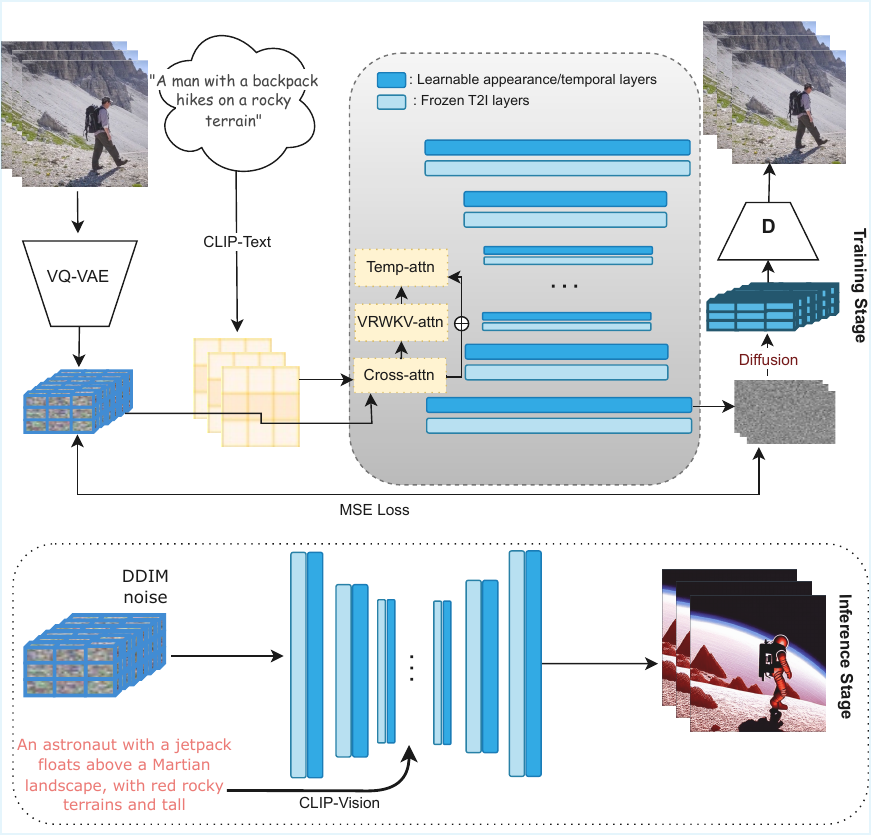}
\caption{Pipeline of \texttt{VRWKV-Editor}: Given a text–video pair (e.g., “A man with a backpack hikes on a rocky terrain”) as input, our method leverages pretrained text-to-image diffusion models for text-to-video generation. The input video is first encoded into a discrete latent space, after which our U-Net architecture predicts the injected noise (a detailed illustration is depicted in Figure~\ref{fig1}). During inference, a novel video is synthesized by inverting the discrete noise from the input video, guided by an edited prompt (e.g., “An astronaut with a jetpack floats above a Martian landscape, with red rocky terrains and tall”).}
    \label{fig2}
\end{figure*}

\section{Method}\label{sec:md}
This section presents our contribution. We first formulate the problem and highlight the computational efficiency issues in existing architectures, then detail the proposed method.
\subsection{Problem Formulation}
Given a video sequence $X = \{X_i\}_1^n$ with $n$ frames, and a text prompt $P$ describing $X$, our goal is to generate an edited video $X^{'} = \{X_i^{'}\}_1^n$ that aligns with new prompts $P^{'}$ provided by the user, where $X_i^{'} \in \mathbb{R}^{C \times H \times W}$ is a frame with C channels, height H and width W.

\subsection{Computational Efficiency Challenges}\label{rpl}
The computational demands of video processing, particularly for high-resolution content and long sequences, present a fundamental bottleneck in current methods \cite{ZHAO2026111867, GAO2026111871}. Traditional ViTs process high-resolution frames resulting in quadratic growth of memory and computation due to the pairwise token interactions in the self-attention mechanism. For example, a 2048$\times$2048 frame (16,384 tokens with a patch size of 16) requires dramatically more memory and processing time than lower-resolution content, making real-time applications impractical. The classical self-attention is computed as follows:

\begin{equation}
\text{Attn}(Q, K, V) = \text{softmax}(QK^\top)V,
\end{equation}

where Q, K and V represents the Querie, Key and Value, respectively. Although effective in capturing long-range dependencies, this operation scales poorly with sequence length.\\
Variants such as AFT \cite{zhai2021attention} replace the dot-product operation in standard attention with element-wise multiplication, leading to a linearized formulation that avoids explicit query–key interactions. Specifically, AFT introduces learnable positional biases $w_{t,i}$ into the exponential weights:
\begin{equation} \text{Attn}^+(W, K, V)_t = \frac{\sum_{i=1}^t e^{w_{t,i} + k_i} \odot v_i}{\sum_{i=1}^t e^{w_{t,i} + k_i}}, \end{equation}
where the terms $w_{t,i}$ encode contextual dependencies and act as learnable position encodings. This design captures structured interactions across tokens while reducing the quadratic cost of standard attention. Nevertheless, even with such improvements, many methods still resort to hierarchical or windowed attention to further reduce memory and computation, at the expense of the global receptive field, which is crucial for modeling long-range dependencies in video understanding. Linear vision transformers, such as \texttt{VRWKV}, address these challenges by reformulating token interactions to achieve linear complexity with respect to sequence length. In RWKV, attention weights are expressed as a channel-wise time decay:
\begin{equation}
w_{t,i} = -(t - i) w, \quad w \in \mathbb{R}_{\ge 0}^d,
\end{equation}
ensuring that contributions from past tokens decay over time. This formulation allows the attention operation to be transformed into a recurrent structure, drastically reducing memory footprint and computational cost. \texttt{VRWKV} achieves up to \textbf{10$\times$ faster inference} and \textbf{80\% memory reduction} compared to traditional ViTs at high resolutions, while maintaining comparable performance. Linear complexity $O(Td)$ is achieved for both forward and backward passes, with further speed gains provided by custom CUDA kernels.
By leveraging linear attention, models such as \texttt{VRWKV} preserve a global receptive field while remaining computationally efficient, making high-resolution, long-sequence video processing feasible for real-time applications.

\subsection{The proposed architecture: \texttt{VRWKV-Editor}}
\begin{figure*}[t]
    \centering
\includegraphics[width=0.98\textwidth]{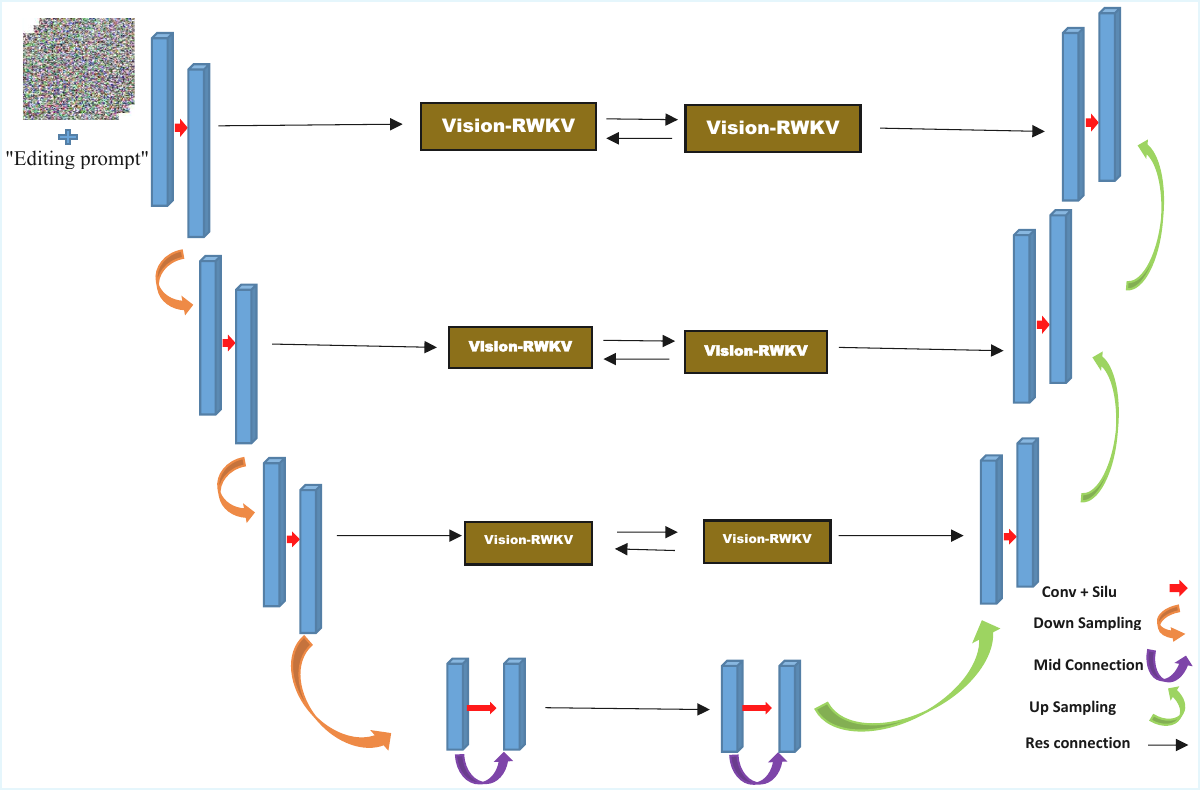}
\caption{Architecture of the U-Net backbone employed in \texttt{VRWKV-Editor}. The design incorporates \texttt{VRWKV} modules within the skip connections, enabling efficient long-range dependency modeling while preserving fine-grained spatial information across down-sampling and up-sampling layers.}
    \label{fig1}
\end{figure*}

Tune-A-Video \cite{wu2023tune} reduced the high computational complexity of spatio-temporal attention (ST-Attn) in generating videos with increasing frames by introducing a novel sparse version of causal attention mechanism, where the attention matrix is computed between frame $X_{i}$ and two previous frames $X_{1}$ and $X_{i-1}$, reducing the complexity from $O((dT)^2)$ to $O(2dT^2)$. However, the quadratic complexity still remains a bottleneck for long sequences of video frames, limiting their ability to efficiently process high-resolution images and lengthy sequences, posing a significant barrier to their broader application. To overcome this issue, we propose a linear attention mechanism based on the \texttt{VRWKV} \cite{duan2024vision} framework, where we replace the traditional quadratic attention computation with a receptance-weighted key-value operation that scales linearly with sequence length, reducing the complexity from $O(2dT^2)$ to $O(2dT)$.
Before aggregation, VRWKV employs a lightweight mixing mechanism to capture local temporal dynamics without heavy convolution. For a given input sequence of video frames $X_t$,  we compute the Receptance $(R_t)$, Key $(K_t)$, and Value $(V_t)$ vectors using a learnable interpolation with the previous time-step state $X_{t-1}$ as defined in Eq.~\ref{eq5}. In the standard Transformer, the softmax prevents the decomposition of the attention matrix. VRWKV replaces this with a channel-wise exponential decay (see Eq.\ref{eq6}). By defining the accumulated states $A_t$ and $B_t$: 
\begin{equation}
A_t=\sum_{i=0}^t e^{-(t-i) w+k_i} v_i, \quad B_t=\sum_{i=0}^t e^{-(t-i) w+k_i}
\end{equation}
We can express the state update recursively:
\begin{equation}
\begin{aligned}
& A_t=e^{-w} \odot A_{t-1}+e^{k_t} \odot v_t \\
& B_t=e^{-w} \odot B_{t-1}+e^{k_t}
\end{aligned}
\end{equation}
This recursive formulation demonstrates that calculating the output for step $t$ depends only on the fixed-size hidden states $A_{t-1}$ and $B_{t-1}$, not on the entire history of previous frames. Consequently, the complexity collapses from $O(2dT^2)$ to $O(2dT)$.\\
Our approach leverages the \texttt{VRWKV's} inherent ability to capture long-range dependencies through its recurrent formulation while maintaining spatial awareness through learnable shift operations and channel mixing.\\
An overview of our approach is depicted in Figure \ref{fig2}. Given a text-video pair as input, our model leverages the pretrained T2I diffusion models for T2V generation by encoding the input video into a discrete space, then we predict the added noise using our U-Net architecture enhanced with \texttt{VRWKV} modules (Figure \ref{fig1}). During inference, we sample a novel video from the discrete noise inverted from the input video, guided by an edited prompt.

\subsection{\texttt{3D-VRWKV} Module}
\begin{figure*}[t]
    \centering
    \includegraphics[width=0.9\linewidth]{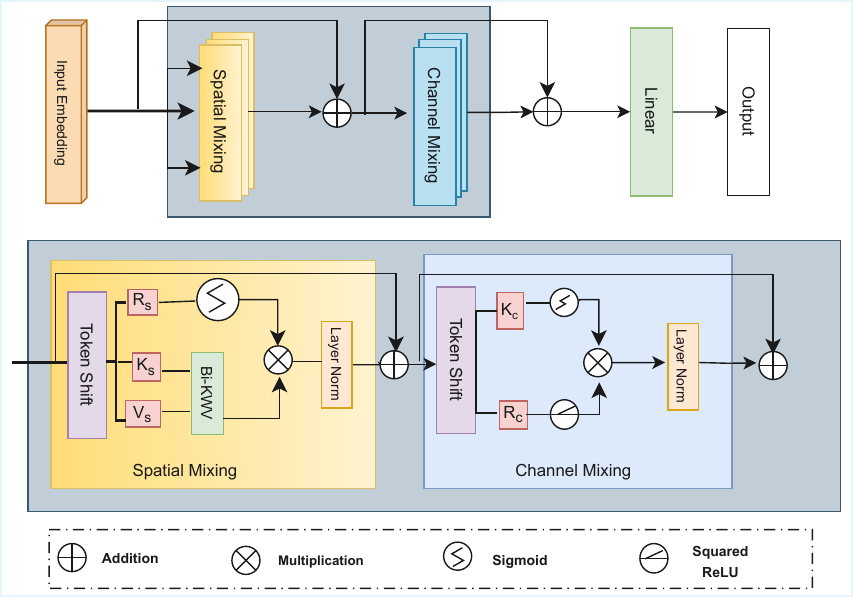}
    \caption{\texttt{VRWKV} \cite{duan2024vision} module employed on \texttt{VRWKV-Editor}. The \texttt{3D-VRWKV} architecture includes identical
\texttt{VRWKV} encoder layer, an average pooling layer, and a linear prediction head. Token Shift denotes the quad-directional shift method tailed for vision tasks.}
    \label{fig:placeholder}
\end{figure*}
 During our experiments, we observed that traditional CNN-based spatiotemporal modules failed to capture complex spatiotemporal dependencies, while adding more cross-attentions or temporal attentions \cite{fan2021multiscale, arnab2021vivit} only increased the complexity of the model without improving the quality of edited videos or maintaining their consistency. Inspired by the RWKV architecture's ability to combine the strengths of RNNs and Transformers, we introduce \texttt{3D-VRWKV} (Figure \ref{fig:placeholder}), a variant of \texttt{VRWKV} module specifically designed for video spatiotemporal modeling.
The 3D-VRWKV module consists of two sequential sub-modules: spatiotemporal mixing and channel mixing, each operating with linear complexity.\\
\textbf{Spatiotemporal Mixing with Bidirectional Processing:}
The spatiotemporal mixing module processes input video features $X$ through temporal interpolation, spatial projection and Bidirectional Weighted Key-Value Attention.\\
To enable the model to learn smooth temporal transitions, we first mix each frame's features with those of its predecessor:

\begin{equation}
X^{mix}_t = \mu \odot X_t + (1 - \mu) \odot X_{t-1}, \quad t=1,...,T 
\end{equation}

where $\mu \in \mathbb{R}^ {1\times1\times1\times1\times C}$ is a learnable channel-wise interpolation parameter.\\
For the spatial projection, the mixed features are projected to receptance, key, and value spaces through learned transformations:

\begin{align}
R_s &= W_R \star X_t^{\text{mix}} \in \mathbb{R}^{B \times H \times W \times d} \\
K_s &= W_K \star X_t^{\text{mix}} \in \mathbb{R}^{B \times H \times W \times d} \\
V_s &= W_V \star {X}_t^{\text{mix}} \in \mathbb{R}^{B \times H \times W \times d}
\end{align}

where $\star$ denotes spatial convolution (implemented as 1×1 convolutions for computational efficiency), and $d = H_{\text{heads}} \times d_{\text{head}}$ is the total embedding dimension. We reshape these projections to separate the head dimension: $R_s, K_s, V_s \in \mathbb{R}^{B \times T \times N \times H_{\text{heads}} \times d_{\text{head}}}$ where $N = H \times W$ is the total number of spatial tokens.\\
\textbf{Channel Mixing:}
After spatiotemporal mixing, we apply a channel mixing module to enable communication across feature channels, which is crucial for modeling correlations between different semantic attributes such as object identity, color, and texture. Similar to the spatiotemporal mixing stage, we begin by creating temporally interpolated channel features through: 
\begin{equation}
Y^{cmix}_t = \mu_c \odot Y^{spatial}_t + (1 - \mu_c) \odot Y^{spatial}_{t-1}
\end{equation}

where $\mu_c \in \mathbb{R}^ {1\times1\times1\times1\times C}$  is a separate learnable parameter for channel mixing that allows the model to adaptively control temporal smoothness in the channel domain. These interpolated features are then projected to receptance and key spaces through learned transformations:

\begin{align}
R_t^c &= W_R^c \star Y_t^{\text{cmix}} \\
{K}_t^c &= {W}_K^c \star {Y}_t^{\text{cmix}}
\end{align}

where $\star$ denotes spatial convolution implemented as $1 \times 1$ convolutions.\\
The channel mixing output is then computed through a squared ReLU activation that provides enhanced non-linearity for modeling complex channel interactions: 
\begin{equation}
\text{CM}_t = \sigma(R^c_t) \odot \max(K^c_t \cdot W_c, 0)^2
\end{equation}
where $W_c$ is an additional learnable transformation matrix and the squared activation $\max(\cdot,0)^2$ provides stronger non-linearity than standard ReLU. The final output combines both spatiotemporal and channel mixing through residual connections:
\begin{equation}
Z_t = X_t + Y^{\text{spatial}}_t + \text{CM}_t
\end{equation}

The \texttt{3D-VRWKV} module addresses key limitations of traditional attention mechanisms in video editing. Unlike quadratic attention mechanisms, \texttt{3D-VRWKV} operates with linear complexity $O(d)$, making it more efficient for processing long video sequences (See Table \ref{tab:sequence_performance}). The module maintains a recurrent state that captures temporal dependencies across frames while preserving spatial relationships. Moreover, the weighted key-value mechanism allows for dynamic focus on relevant spatiotemporal features.
\begin{figure*}[tbp]
    \centering
\includegraphics[scale=0.65]{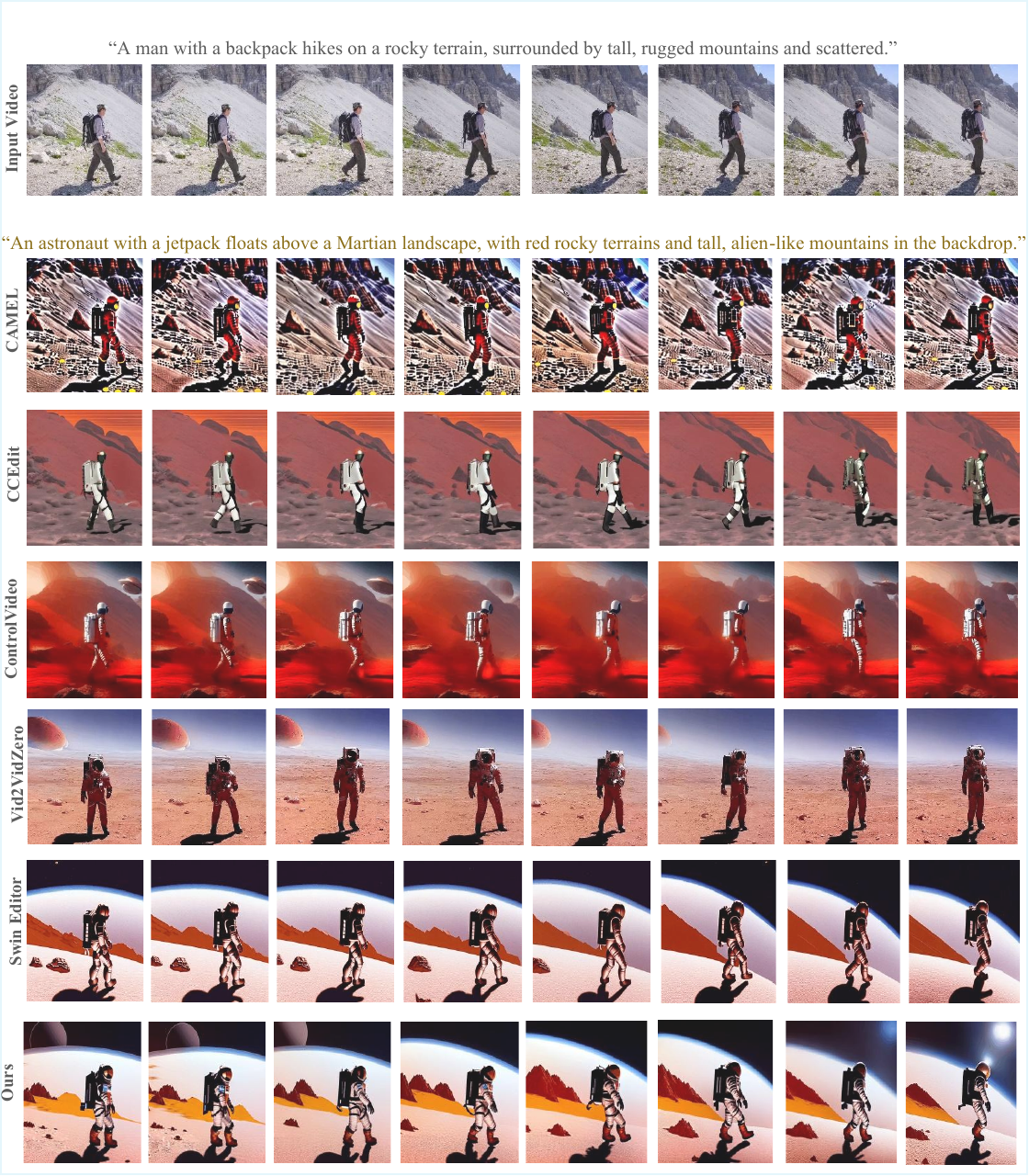}
\caption{Qualitative comparison between CAMEL \cite{Zhang_2024_CVPR}, CCEdit \cite{Feng_2024_CVPR}, ControlVideo \cite{zhang2023controlvideo}, Vid2Vid-zero \cite{wang2023zero}, Swin-Editor \cite{aitrouga2025swin} and our method. The first row presents the input video. The figure highlights two different scenarios: Object editing and object-background editing . The results showcase the superiority of our framework in editing videos across various scenarios, producing consistent and high quality outputs.
}
    \label{fig3}
\end{figure*}

\section{Experiments}\label{sec:exp}
We conduct experiments to test the performance of \texttt{VRWKV-Editor} with existing video editing methods. Our evaluation encompasses computational efficiency metrics, generation quality assessments, and temporal coherence analysis across various video editing tasks. 

\begin{table}[t]
\centering
\footnotesize
\caption{Quantitative comparison between our method and evaluated baselines in terms of frame consistency, textual alignment, and runtime for overall pipelines (minutes).}
\label{Tab1}
\begin{tabularx}{\linewidth}{l *{6}{>{\centering\arraybackslash}X}}
\toprule
\multirow{2}{*}{\textbf{Method}} 
  & \multicolumn{2}{c}{\textbf{Frame Consistency}} 
  & \multicolumn{2}{c}{\textbf{Textual Alignment}} 
  & \multicolumn{2}{c}{\textbf{Runtime [min]}} \\
\cmidrule(lr){2-3} \cmidrule(lr){4-5} \cmidrule(lr){6-7}
& {\makecell{Clip \\ Score $\uparrow$}} 
& {\makecell{User \\ Vote $\uparrow$}} 
& {\makecell{Pick \\ Score $\uparrow$}} 
& {\makecell{User \\ Vote $\uparrow$}} 
& {\makecell{Training $\downarrow$}} 
& {\makecell{Inference $\downarrow$}} \\
\midrule
CAMEL        & 93.38 & 14.20 & 26.67 & 8.60  & 13.00 & 1.50 \\
CCEdit       & 93.07 & 10.10 & 25.70 & 11.20 & 14.50 & 2.00 \\
Tune-A-Video & 95.22 & 17.30 & \textbf{28.18} & 20.10 & 11.00 & 0.50 \\
Video-P2P    & \textbf{96.40} & 11.60 & 25.82 & 8.10  & 22.00 & 2.00 \\
Vid2Vid-zero & 94.18 & 8.60  & 26.98 & 14.00 & \textbf{3.00}  & 3.00 \\
Swin-Editor  & 95.47 & 18.40 & 28.05 & 18.70 & 10.00 & 0.30 \\
\midrule
\texttt{\textbf{VRWKV-Editor}} & \textbf{95.75} & \textbf{19.80} & \textbf{27.90} & \textbf{19.30} & \textbf{8.00} & \textbf{0.16} \\
\bottomrule
\end{tabularx}
\end{table}

\subsection{Implementation Details}
For VRWKV, we implemented a single layer of RWKV-V4 (Figure \ref{fig2}) with local initialization mode, where the spatial mixing parameters are initialized to 1.0, making it behave more like local attention initially. The number of attention heads is set to 16 with 88 dimension for each head (total embedding dimension of 1408). The vision-specific hyperparameters include patch size of 16×16, input frame resolution of 512×512, spatial shift pixel of 1, and channel gamma of 1/4 for directional shifting across four spatial directions (up, down, left, right), enabling efficient processing of high-resolution frames without windowing operations.\\
We used the pretrained weights of Stable Diffusion v1.5 \cite{rombach2022high} to initialize the 2D models of our framework.  In the training stage, the AdamW optimizer is employed with a learning rate of $3e^{-5}$ for a total of 300 to 500 steps with 100 validation steps and a batch size of one video. For each video, we designed 3 edited prompts covering objects, background, color, weather, or style changing. During inference, we implemented the DDIM sampler \cite{saharia2022photorealistic} alongside classifier-free guidance \cite{ho2022classifier} of $7.5$, controlling how strongly the model follows the text prompt. All experiments are conducted on a single machine with an NVIDIA A100-SXM4-80GB GPU.

\begin{figure}[t]
    \centering
\includegraphics[scale=0.4]{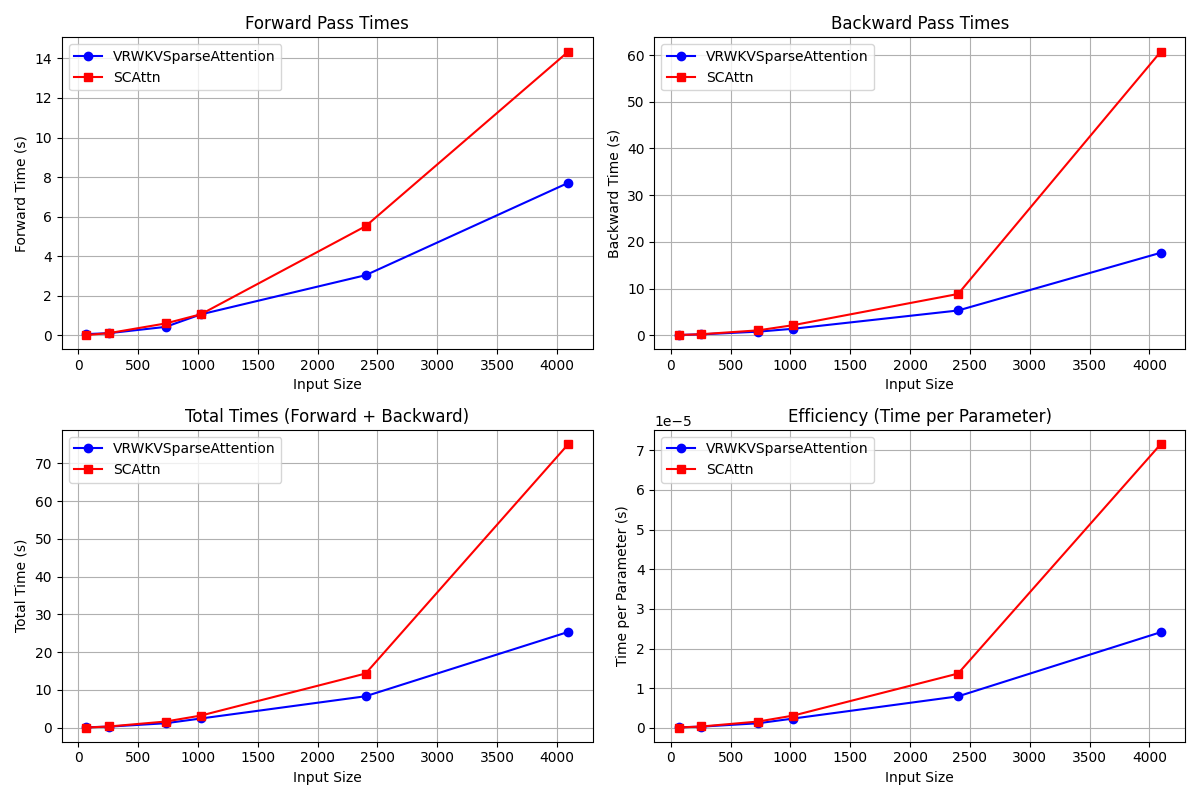}
\caption{Comparison of runtime between our \texttt{VRWKV} Sparse Attention (as deployed in our model) and the Sparse Causal Attention used in Tune-A-Video \cite{wu2023tune} is presented, showing forward pass, backward pass, and total execution times across varying input sizes. The execution time for \texttt{VRWKV} Sparse Attention scales linearly with input size, whereas the Sparse Causal Attention in Tune-A-Video exhibits quadratic growth.}
    \label{fig5}
\end{figure}
\begin{figure}[t]
    \centering
\includegraphics[scale=0.4]{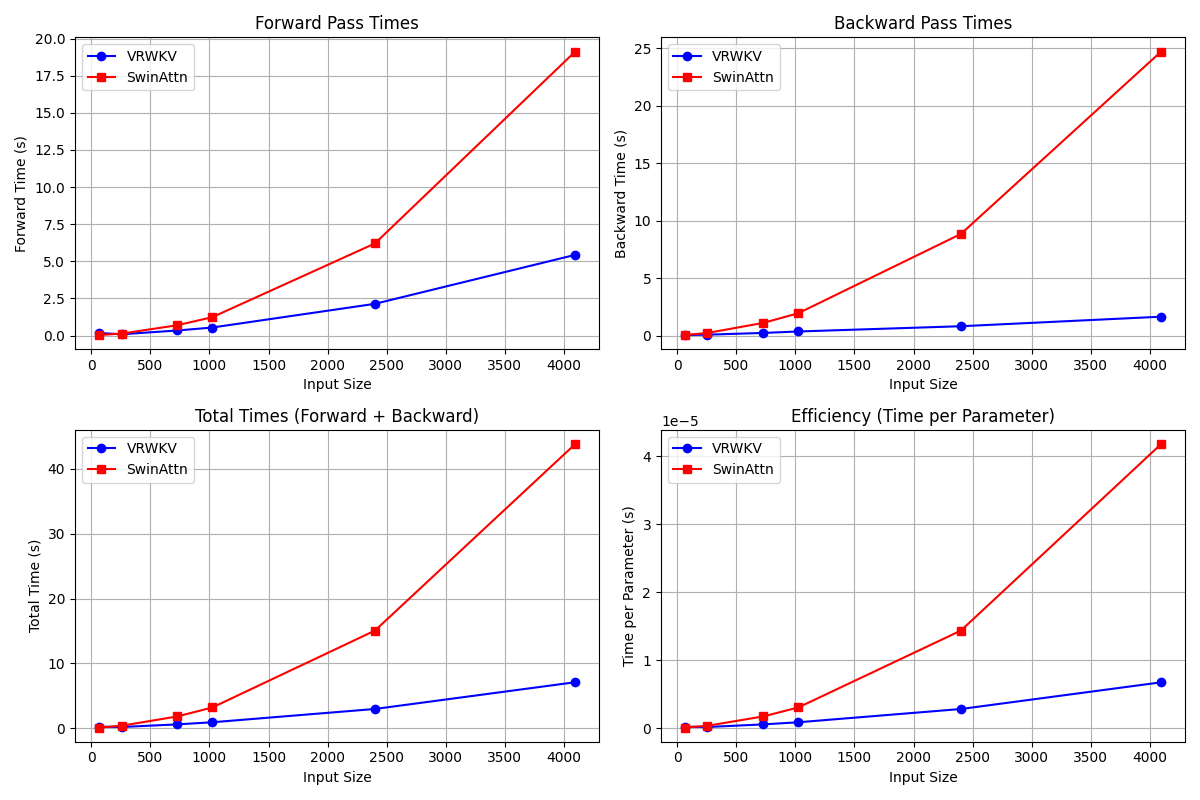}
\caption{Computational timing comparison between \texttt{VRWKV} Attention and Swin Attention deployed in our model showing forward pass, backward pass, and total execution times across different input sizes. \texttt{VRWKV-Editor} demonstrates increasing efficiency gains with larger inputs, achieving up to 3.68× speedup for 4096×4096 inputs.}
    \label{fig4}
\end{figure}

\textbf{Dataset and Metrics:} To evaluate our approach, we extended the Video-P2P dataset \cite{liu2023video} by incorporating 13 additional videos from the DAVIS dataset \cite{ponttuset20182017davischallengevideo}, resulting in a total of 24 videos. Each video was paired with three text prompts, yielding 72 edited samples overall. For evaluation, we report CLIP-Score and Pick-Score to measure semantic alignment and perceptual quality, respectively, alongside user preference studies and runtime analysis.

\textbf{Evaluated Models:} We compared our method against six recent baselines:
CAMEL \cite{Zhang_2024_CVPR}, CCEdit \cite{Feng_2024_CVPR}, Tune-A-Video \cite{wu2023tune}, Video-P2P \cite{liu2023video}, Vid2Vid-zero \cite{wang2023zero} and Swin-Editor \cite{aitrouga2025swin}. Experiments are conducted with their official codes and configurations.
\subsection{Computational Efficiency Analysis:}
\textbf{Timing Performance Comparison:}
Our timing experiments reveal significant computational advantages of \texttt{VRWKV-Editor} over window-based attention architectures. Figures \ref{fig5} and \ref{fig4} presents the forward and backward pass timing results for different input sequence lengths, and running time of end-to-end editing pipelines for all evaluated methods are reported in Table \ref{Tab1}.
The results demonstrate that \texttt{VRWKV-Editor} achieves substantial speedups that increase with input size, confirming the theoretical $O(n)$ vs $O(n^2)$ complexity advantage. For large inputs (4096×4096), our method achieves up to 3.68× speedup in total computation time.

\textbf{Quantitative Results:} In Table \ref{Tab1}, we report the quantitative results of \texttt{VRWKV-Editor} versus state-of-the-art video editing methods. As the results shows, our model achieves a trade-off between temporal consistency, text alignment
(for objects, background, and style) and high-quality video editing while significantly reducing computational complexity. This is also evidenced by the performance obtained in both automatic evaluations and user studies.

\textbf{Qualitative Evaluation:} Figure \ref{fig2} showcases a visual comparison between our method and the other baselines. Further results are shown in Figure \ref{fig0} and on the project page \href{https://abdo-rg.github.io/VRWKV-Editor/}{\texttt{VRWKV-Editor}}. We conducted our evaluation by editing objects, background, style or combination between them.

\begin{table}[htbp]
\centering
\caption{Performance according to Video Length.}
\label{tab:sequence_performance}
\begin{tabularx}{\linewidth}{c *{3}{>{\centering\arraybackslash}X}}
\toprule
\textbf{Video Length} & \textbf{Swin-Attn Time (s)} & \textbf{\texttt{VRWKV}-Attn Time (s)} & \textbf{Memory Ratio} \\
\midrule
16 frames  & 2.01 & 1.02 & 1.97× \\
32 frames  & 3.99 & 1.78 & 2.24× \\
64 frames  & 7.95 & 3.54 & 2.24× \\
128 frames & 15.89 & 7.05 & 2.25× \\
\bottomrule
\end{tabularx}
\end{table}

\subsection{Hyperparameter Robustness}

To evaluate the robustness of \texttt{VRWKV-Editor}, we assess the sensitivity to three key architectural hyperparameters: the spatial shift pixel, the channel gamma controlling the proportion of shifted channels, and the number of attention heads. Each hyperparameter is varied independently while keeping the remaining settings fixed at their default configuration.

\begin{figure}[t]
    \centering
    \includegraphics[width=0.85\linewidth]{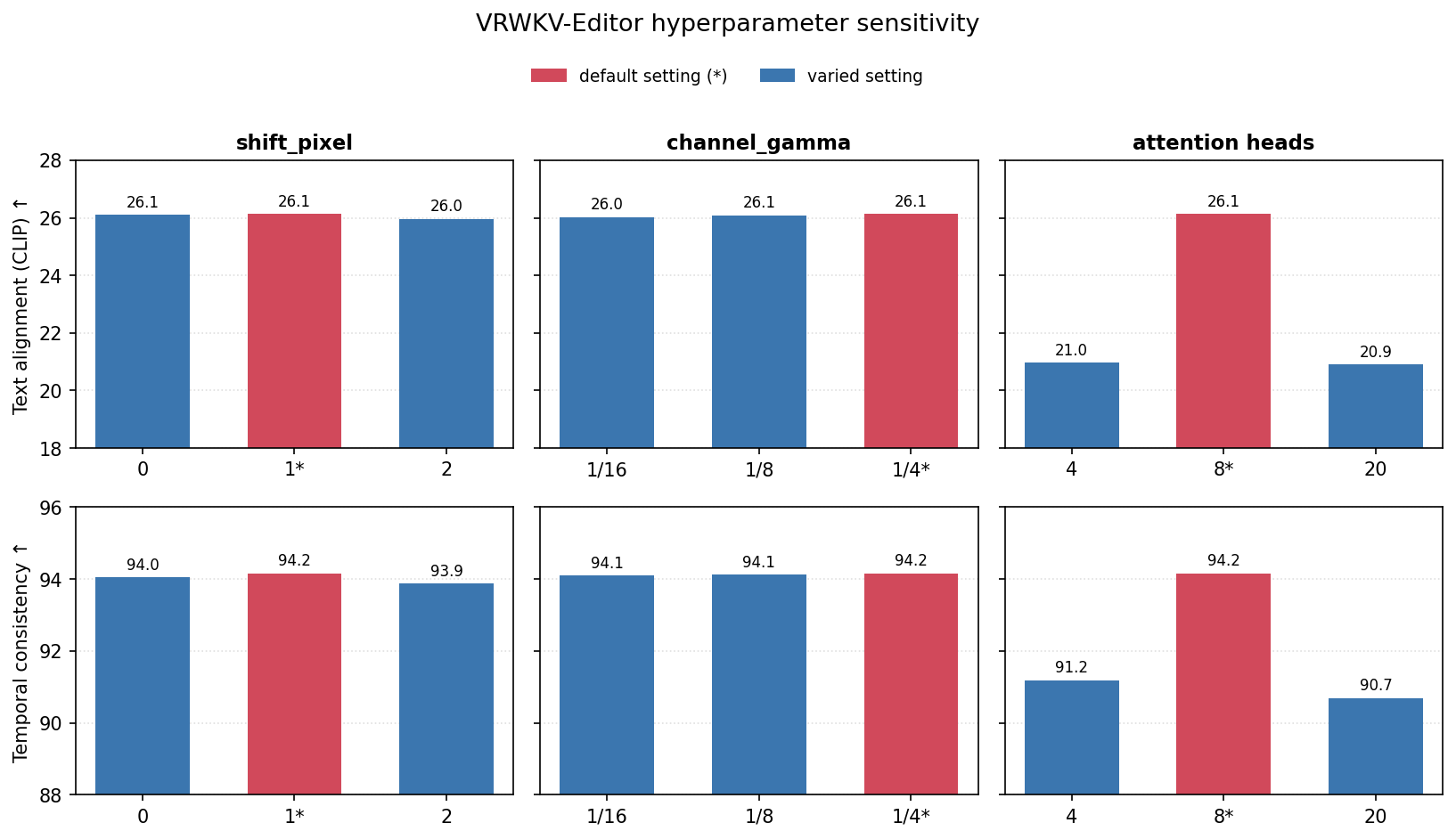}
    \caption{Hyperparameter robustness of \texttt{VRWKV-Editor}. The default configuration is highlighted in each panel. The proposed method is largely insensitive to the spatial shift pixel and channel gamma, with only minor variations in text alignment and temporal consistency. In contrast, the number of attention heads has a more pronounced influence, with the default setting of eight heads consistently achieving the best performance.}
    \label{fig:hyperparameter_robustness}
\end{figure}

Figure~\ref{fig:hyperparameter_robustness} shows that the proposed architecture is robust to the spatial mixing hyperparameters. Varying the spatial shift pixel or channel gamma produces only negligible changes in both CLIP text-alignment and temporal consistency, indicating that the performance of \texttt{VRWKV-Editor} is stable across a broad range of spatial configurations. The number of attention heads has a larger influence on editing quality, with the default setting of eight heads consistently yielding the best results among the evaluated configurations. We attribute this to our pretrained initialization strategy: eight heads preserve the per-head feature dimensionality expected by the pretrained Stable Diffusion weights \cite{rombach2022high}, whereas changing the head count alters the feature partitioning and degrades the effectiveness of weight transfer.

\subsection{Ablation Studies}
\subsubsection{Component-wise Analysis} Table~\ref{tab:ablation_study} summarizes the impact of our design choices on textual alignment, temporal consistency, and forward pass time. Starting from the Swin-Editor baseline, replacing only the attention mechanism with \texttt{VRWKV} attention reduces the forward time from 30\,s to 18\,s with only a minor change in editing performance. Replacing the trainable CrossAttention (Full SCAttn) further improves textual alignment and temporal consistency while reducing the forward time to 14\,s. Incorporating the complete \texttt{VRWKV-Editor} design yields the highest temporal consistency (95.75) while maintaining competitive textual alignment (27.90) and a low forward time (16\,s). Overall, the proposed model provides the best balance between editing quality and computational efficiency.
\begin{table}[htbp]
\centering
\footnotesize
\caption{Ablation study results on textual alignment, temporal consistency, and inference speed. The proposed \texttt{VRWKV-Editor} achieves the best balance between accuracy and efficiency.}
\label{tab:ablation_study}
\begin{tabularx}{\linewidth}{l *{3}{>{\centering\arraybackslash}X}}
\toprule
\textbf{Configuration} & \textbf{Textual alignment} $\uparrow$ & \textbf{Temp. Consistency} $\uparrow$ & \textbf{Forward Time (s)} $\downarrow$ \\
\midrule
Swin-Editor (baseline)   & 28.05 & 95.47 & 30 \\
+ \texttt{VRWKV} attention only   & 27.87 & 95.68 & 18 \\
+ Full SCAttn            & 28.20 & 95.73 & 14 \\
+ Full VRWKV-Editor      & \textbf{27.90} & \textbf{95.75} & \textbf{16} \\
\bottomrule
\end{tabularx}
\end{table}

\subsubsection{Scalability with Increasing Sequence Length}
Table \ref{tab:sequence_performance} demonstrates the scalability of different video editing methods across varying video lengths. As the sequence length increases from 16 to 128 frames, the computational advantages of linear attention mechanisms become increasingly pronounced. Methods based on quadratic attention mechanisms show substantial runtime growth, with Video-P2P's inference time increasing from 2.00 to 17.00 minutes and Vid2Vid-zero growing from 3.00 to 24.00 minutes. Most notably, Swin-Editor exhibits the steepest scaling curve, escalating from 0.30 to 19.20 minutes ($64\times$ increase) due to its window-based attention mechanism's sensitivity to long sequences. In contrast, \texttt{VRWKV-Editor} scales linearly and efficiently, requiring only 1.28 minutes for 128 frames while maintaining superior performance metrics.

\begin{table}[htbp]
\centering
\footnotesize
\caption{Estimated runtime scaling (in minutes) of all evaluated methods across different video lengths. 
Runtimes are extrapolated from the official pipeline runtimes in Table~\ref{Tab1} based on each method's temporal processing complexity. Bold indicates best performance, underline indicates second-best.}
\label{tab:sequence_performance}
\begin{tabularx}{\linewidth}{l *{4}{>{\centering\arraybackslash}X}}
\toprule
\textbf{Method} & \textbf{16 frames} & \textbf{32 frames} & \textbf{64 frames} & \textbf{128 frames} \\
\midrule
CAMEL        & 1.50 & 3.00 & 6.00 & 12.00 \\
CCEdit       & 2.00 & 4.00 & 8.00 & 16.00 \\
Tune-A-Video & 0.50 & 1.22 & \underline{2.80} & \underline{6.50} \\
Video-P2P    & 2.00 & 4.00 & 8.50 & 17.00 \\
Vid2Vid-zero & 3.00 & 6.00 & 12.00 & 24.00 \\
Swin-Editor  & \underline{0.30} & \underline{1.20} & 4.80 & 19.20 \\
\midrule
\texttt{\textbf{VRWKV-Editor}} & \textbf{0.16} & \textbf{0.32} & \textbf{0.64} & \textbf{1.28} \\
\bottomrule
\end{tabularx}
\end{table}

\subsubsection{Ablation of channel-mixing activation functions and learned mixing parameters}

\begin{table}[t]
\centering
\caption{Ablation of channel-mixing activation functions. Squared-ReLU achieves the best overall performance while maintaining comparable inference time.}
\label{tab:activation_ablation}

\begin{tabularx}{\linewidth}{l >{\centering\arraybackslash}X >{\centering\arraybackslash}X >{\centering\arraybackslash}X}
\toprule
\textbf{Activation} &
\textbf{Text Align.} &
\textbf{Temporal Cons.} &
\textbf{Time (s)} \\
\midrule
No Channel Mixing & 25.19 & 91.62 & 0.283 \\
ReLU             & 25.89 & 90.77 & 0.281 \\
GELU             & 26.21 & 90.95 & 0.281 \\
Swish            & 26.43 & 90.48 & 0.281 \\
Squared-ReLU     & \textbf{28.96} & \textbf{93.28} & \textbf{0.263} \\
\bottomrule
\end{tabularx}
\end{table}

To justify the choice of the channel-mixing activation, we compare Squared-ReLU against ReLU, GELU, and Swish. As shown in
Table~\ref{tab:activation_ablation}, all channel-mixing variants
perform competitively, while Squared-ReLU achieves the best overall text alignment and temporal consistency with slightly lower inference time. These results support its adoption in the final \texttt{VRWKV-Editor} architecture.

\begin{table}[t]
\centering
\caption{Statistics of the learned interpolation parameters after training. Both parameters deviate substantially from their initialization, indicating adaptive spatio-temporal and channel mixing.}
\label{tab:mixing_parameters}

\begin{tabularx}{\linewidth}{l >{\centering\arraybackslash}X >{\centering\arraybackslash}X >{\centering\arraybackslash}X >{\centering\arraybackslash}X}
\toprule
\textbf{Parameter} &
\textbf{Mean} &
\textbf{Std.} &
\textbf{Range} &
\textbf{Mean $|x-0.5|$} \\
\midrule
$\mu$   & 0.593 & 0.385 & $[-0.87,\;2.24]$ & 0.256 \\
$\mu_c$ & 0.531 & 0.515 & $[-1.48,\;2.57]$ & 0.335 \\
\bottomrule
\end{tabularx}
\end{table}

To better understand the behavior of the learnable interpolation
parameters, we analyze their statistics after training.
Table~\ref{tab:mixing_parameters} shows that both $\mu$ and $\mu_c$ exhibit broad learned distributions and deviate substantially from their initialization. In particular, values extending beyond the interval $[0,1]$ indicate that the model is not restricted to fixed convex interpolation, but instead learns to adaptively amplify or attenuate shifted features according to the input.

\section{Limitations: }\label{sec:lm}
A key conceptual limitation of the current \texttt{VRWKV-Editor} lies in its use of \texttt{VRWKV} in a Transformer-style mode, without returning recurrent states across time. While this formulation aligns with conventional transformer architectures, it does not leverage \texttt{VRWKV’s} native RNN mode, which enables true sequential processing with memory.\\
Adapting \texttt{VRWKV-Editor} to operate in its RNN mode presents a promising research direction. By returning and propagating states during inference, the model could process arbitrarily long videos chunk by chunk while maintaining temporal coherence through memory linking. This would exploit the fundamental advantage of RNNs over transformers, which require the full sequence as input or resort to independent chunking, thereby limiting scalability. Such an extension could unlock \texttt{VRWKV’s} potential for efficient editing of very large-scale videos.\\
Finally, our evaluation relies primarily on CLIP-based scores and user studies, which, while useful, do not fully capture video-specific qualities such as temporal consistency and motion smoothness. Future work should therefore prioritize adapting VRWKV-Editor to its RNN mode for scalable long-sequence editing, and adopting more sophisticated video-centric evaluation metrics.

\section{Conclusion:}\label{sec:cc}
In this paper, we presented \texttt{VRWKV-Editor}, a novel video editing framework that leverages \texttt{VRWKV's} linear spatio-temporal attention mechanism to address the computational limitations of traditional quadratic attention methods. The exploration of \texttt{VRWKV} in video editing task demonstrates its potential to be a viable alternative to traditional video understanding models, successfully reducing computational complexity while maintaining high-quality editing performance. The successful integration of RWKV architectures into video editing workflows opens new possibilities for efficient spatiotemporal modeling and real-time video processing applications. Future work should focus on optimizing low-level GPU kernels and extending the framework to broader video understanding tasks. This research contributes to the advancement of efficient video processing architectures and demonstrates the potential of linear attention mechanisms in computer vision tasks.

\bibliographystyle{unsrt}  
\bibliography{references}  


\end{document}